# INCORPORATING GRANULARITY BIAS AS THE MARGIN INTO CONTRASTIVE LOSS FOR VIDEO CAPTIONING

GU JIAYANG[1], YAO FENGMING[2]

[1]The School of Computer Science and Engineering, University of Electronic Science and Technology of China, Chengdu 611731, China
[2]The School of Aeronautics and Astronautics, University of Electronic Science and Technology of China, Chengdu 611731, China
E-MAIL: jiayang.barrygu@gmail.com, yaofengming9901@foxmail.com

**Abstract:**

Video captioning models easily suffer from long-tail distribution of phrases, which makes captioning models prone to generate vague sentences instead of accurate ones. However, existing debiasing strategies tend to export external knowledge to build dependency trees of words or refine frequency distribution by complex losses and extra input features, which lack interpretability and are hard to train. To mitigate the impact of granularity bias on the model, we introduced a statistical-based bias extractor. This extractor quantifies the information content within sentences and videos, providing an estimate of the likelihood that a video-sentence pair is affected by granularity bias. Furthermore, with the growing trend of integrating contrastive learning methods into video captioning tasks, we use a bidirectional triplet loss to get more negative samples in a batch. Subsequently, we incorporate the margin score into the contrastive learning loss, establishing distinct training objectives for head and tail sentences. This approach facilitates the model's training effectiveness on tail samples. Our simple yet effective loss, incorporating Granularity bias, is referred to as the Margin-Contrastive Loss (GMC Loss). The proposed model demonstrates state-of-the-art performance on MSRVTT with a CIDEr of 57.17, and MSVD, where CIDEr reaches up to 138.68.

## 1. Introduction

Video Captioning[1,2] aims to generate an accurate and comprehensive sentence corresponding to its given video. It has broader application prospects in fields such as automatic interpretation, navigation assistance, and intelligent human-machine environment development. However, due to the spatiotemporal characteristics and reliance of the generalization ability of model, it is more challenging than any other multimodal downstream task.

As per an intuitive cognition: observing distinct segments of a video engenders varying levels of comprehension regarding the video's content. Notably, the dataset annotations for video captioning show obvious fluctuations in informational content and descriptive misalignment. For instance, consider the difference between "a dog drinks from a swimming pool" and "a dog is standing on the steps of a swimming pool playing with the foamy water created by a garden hose." Despite both sentences delineating the content of a video featuring a common subject, namely a dog and water, the divergent temporal focus and biased linguistic preferences of annotators contribute to discernible disparities in the **granularity** of sentence descriptions. We demonstrate this problem in Fig. 1 by calculating CIDEr for both sentences and videos.

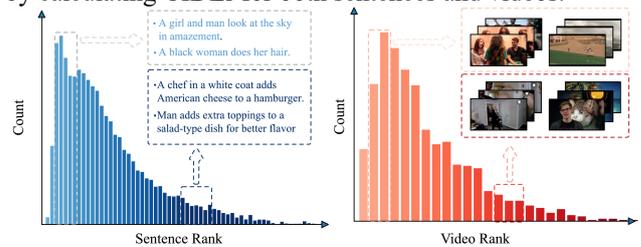

Fig. 1 A visualization of frequency distribution of information content for sentences and videos in the MSRVTT dataset. The horizontal axis represents the transformed rank of the score for sentences or videos, where a higher rank corresponds to greater information content. The vertical axis indicates the frequency of occurrence of each rank in the training set.

Existing methodologies typically address this bias through two principal approaches. 1) *Pre-training model on web-scale dataset.* Leveraging large-scale pre-training models[3,4,5] trained on web-scale datasets[6,7] can instill a more uniformly acknowledged corpus within the model, which aspires inference model to cultivate a heightened awareness of a more standardized linguistic convention. 2) *Extra annotation of visual and textual alignment*. Mitigating this misalignment involves the incorporation of external features and supplementary annotation information. This facilitates the alignment of video and textual modalities across distinct granularities. During the predictive phase, structured generative rules are applied to constrain the model from outputting certain canonical terms.

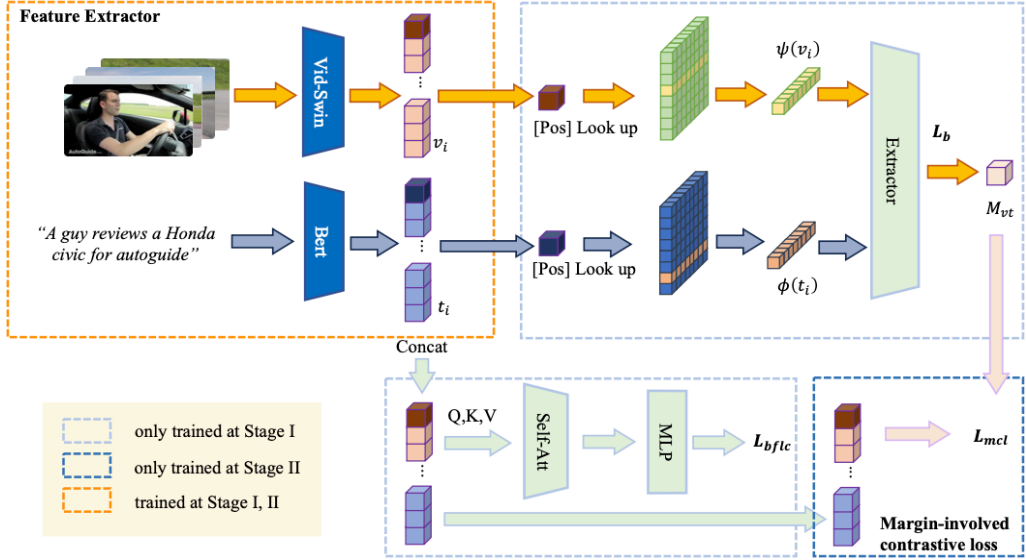

Fig. 2 Framework of our model. It consists of three main blocks: 1) Information content bias extractor; 2) Bidirectional fusion contrastive loss; 3) Adaptive margin-involved contrastive loss.

Concerning the aforementioned issue, we have an intuitive hypothesis: sentences with higher information content occur less frequently in the dataset. Aligned with this premise, we introduce an information content extractor that computes the corresponding information content for both sentences and videos. This score, applicable to any video-sentence pair, signifies the likelihood of encapsulating a greater amount of information. A lower score indicates a higher probability that the video-sentence pair resides in the tail entities of the distribution. Upon the completion of the training for this score, we incorporate it as an additional angle for positive samples in the triplet loss of contrastive learning. This augmentation is integrated into an enhanced *bidirectional fusion contrastive learning framework*, applied during the training of a batch of video-sentence pairs for contrastive learning. Through this training methodology, we not only bidirectionally enhance the training efficacy for tail-end videos and sentences but also refrain from compromising the training effectiveness for head-end videos and sentences. This augmentation results in an overall improvement in training efficacy.

## 2. Proposed Method

In this section, we present an overview of our proposed method. As shown in Fig. 2, it consists of three blocks: 1) Information content bias extractor; 2) Bidirectional fusion contrastive loss; 3) Adaptive margin-involved contrastive loss.

### 2.1. Information Content Bias Extractor

Before mitigating the impact of varying information content at the heads and tails of video-sentence pairs on model training, we are trying to quantify the likelihood of each video-sentence pair being affected by the long-tail problem. A straightforward approach involves a direct comparison of evaluation results between balanced and unbalanced pairs. However, this method is susceptible to inductive bias, wherein prediction errors accumulate throughout each training iteration, proving difficult to eliminate. In contrast, a statistically grounded evaluation metric is a more objective alternative. Aligned with the assumption of the CIDEr method, where the fewer occurrences of an n-gram phrase in the entire lexicon, the more specialized its conveyed information, and consequently, the greater the information it imparts. Inspired by this, we compute the CIDEr score for each sentence appearing in the training set in the corpus which is made of all sentences that appear in the training phase.

Following this rationale, we initiate the training of an additional module termed the Information Content Bias Extractor. This extractor comprises two components: the sentence bias extractor and the video bias extractor. For the former, the module takes as input the CIDEr value of the sentence, rounded to two decimal places. As for the latter, the input is the rounded average CIDEr value of all sentences corresponding to the given video, rounded to one decimal place. We abstractly denote the extractor as $\hat{y}_b(i_v, i_t)$, with the following formula:

$$\hat{y}_b(i_v, i_t) = s(\psi(v_i), \phi(t_i)) \doteq cos(\hat{\xi}_{vt}). \quad (1)$$

Here, $\phi(i_t)$ represents the information content encoder for sentences, and $\psi(i_v)$ represents the information content encoder for videos, both mapping CIDEr statistics of sentences and videos to a $d$-dimensional space. $s(,)$ denotes cosine similarity, $\hat{\xi}$ signifies the angle between $i_v$ and $i_t$, and $\mathcal{N}_v$ refers to the in-batch negative sentences.

We then optimize this angle by the following triplet $\mathcal{L}_b$ loss to optimize the bias extractor:

$$-\sum_{(v,t)\in O^+} \log \frac{\exp(\cos(\hat{\xi}_{vt})/\tau_1)}{\exp(\cos(\hat{\xi}_{vt})/\tau_1) + \sum_{j\in\mathcal{N}_v} \exp(\cos(\hat{\xi}_{vj})/\tau_1)}. \quad (2)$$

This triplet loss incentivizes the extractor to adjust the scores learned for each pair based on the frequency of occurrences of video-text pairs. Simultaneously, according to our proposed hypothesis: the lower the frequency, the smaller the granularity of description for the pair, and consequently, the greater the overall information content. It is intuitive to recognize that pairs with smaller granularity in the tail end of the video-text spectrum learn smaller values in this score, and vice versa. In summary, the scores obtained by this extractor can substantially reflect the predictive difficulty of each sentence.

## 2.2. Bidirectional Fusion Contrastive Loss

In this section, we introduce a bidirectional triplet loss, enabling the model to enhance contrastive learning training effectiveness even with a small batch size by constructing additional in-batch negative samples.

As depicted in Fig. 2, we concatenate video features with the textual features of positive and negative samples, inputting them into a self-attention network. Subsequently, they undergo an MLP layer to obtain similarity scores, forming a matrix of size B*B. Following (3), norm function is applied separately to the video and sentence. The $L_{bfcl}$ is as follows:

$$Norm(m_1, m_2) = \log \frac{exp(cos(\hat{\theta}_{m_1 m_2}/\tau_2))}{\sum_{j\in\mathcal{N}_{m_1}} exp(cos(\hat{\theta}_{m_1 m_2}/\tau_2))}, \quad (3)$$

$$L_{bfcl} = -\frac{Norm(v,t) + Norm(t,v)}{2}. \quad (4)$$

Due to the warm up training time for $\hat{y}_b(i_v, i_t)$, we employ the Bidirectional Fusion Contrastive Loss to facilitate the initial phase of training for the spatial representation of videos and sentences. This enables the initialization of the angles between positive and negative samples in the representation space.

## 2.3. Adaptive Margin-involved Contrastive Loss

Following the completion of Bidirectional Fusion Contrastive Loss, initializing the spatial representation of video and text features, we proceed to explore how the trained bias scores can have a discriminative impact on the head and tail sentences. To avoid disrupting the effectiveness of the warm-up training phase, we continue to employ triplet loss as the loss function during this stage. Simultaneously, guided by an intuitive assumption, we incorporate the bias score as an additional angle into the triplet loss for the angle of positive samples. By introducing differentiation targets for head and tail entities, this approach directs the model's attention more explicitly towards improving the training effectiveness on tail samples. The forementioned margin $M_{vt}$ for video-sentence pair (v, t), and $L_{mcl}$ is formulated as:

$$-\sum_{(v,t)\in O^+} \log \frac{\exp(\cos(\hat{\theta}_{vt} + M_{vt})/\tau_3)}{\exp(\cos(\hat{\theta}_{vt} + M_{vt})/\tau_3) + \sum_{j\in\mathcal{N}_v} \exp(\cos(\hat{\theta}_{vj})/\tau_3)}, \quad (5)$$

$$M_{vt} = min\{\hat{\xi}_{vt}, \pi - \hat{\theta}_{vt}\}. \quad (6)$$

$M_{vt}$ serves as the bias angle derived from Equation(1), and $\pi - \hat{\theta}_{vt}$ is designed to ensure that $cos()$ maintains its property of strict monotonic decrease. In terms of rationality, assume that $v_i$ and $s_i$ are positive samples, where $s_i$ represents a larger granularity, containing less information, as a head sentence. Meanwhile, $s_j$ represents a smaller granularity, containing more information, as a tail sentence. Furthermore, $s_k$ denotes a negative sample sentence not matching $v_i$. In the traditional triplet loss, its aim is to optimize $\cos(\hat{\theta}_{ii})$ being the same as $\cos(\hat{\theta}_{ij})$, and far greater than $\cos(\hat{\theta}_{ik})$. Thus, it gives rise to an inequality, specifically, $\hat{\theta}_{ii} = \hat{\theta}_{ij} < \hat{\theta}_{ik}$.

Upon introducing this margin, which takes on different values based on the positions of videos and sentences in the information content distribution, we formulate a new criterion. In the discrimination between positive and negative samples, assuming $\hat{\theta}_{ii} + M_{ii} < \hat{\theta}_{ik}$, allows for a smaller $\hat{\theta}_{ii}$ and consequently, a smaller angle between positive samples. In the discrimination between head and tail samples, the larger $M_{ii}$ for head samples $s_i$ and the smaller $M_{ij}$ for tail samples $s_j$ need to satisfy the triplet loss requirement $\hat{\theta}_{ii} + M_{ii} = \hat{\theta}_{ij} + M_{ij}$. From this, we can infer that $\hat{\theta}_{ij} = \hat{\theta}_{ii} - \Delta M$, where $\Delta M$ is a positive value (as $\Delta M = M_{ii} - M_{ij}$). Consequently, the target angle for $\hat{\theta}_{ij}$ becomes smaller than that for $\hat{\theta}_{ii}$, thereby achieving a better fitting effect for tail sentences.

It is noteworthy that we have not discarded the loss originally employed for the captioning task. We utilized the caption training loss from the selected baseline[8], denoted as $L_{gen}$. The total loss function is written as the combination of the aforementioned loss as:

$$L_{gmc} = L_{gen} + L_{bfcl} + L_b + L_{mcl} \qquad (7)$$

## 3. Experiment

### 3.1. Experimental Settings

**Dataset.** We conduct extensive experiments on MSR-VTT[9] and MSVD[10] video-text datasets. The MSR-VTT dataset comprises 10,000 video clips, each accompanied by approximately 20 descriptive sentences. This dataset employs 257 textual queries to search for relevant YouTube videos for data collection. Utilizing the official splits, the training set encompasses 6,573 videos, the validation set comprises 497 videos, and the test set contains 2,990 videos. The MSVD dataset consists of 1,970 videos, with each video corresponding to approximately 40 sentences. In accordance with the partitioning method outlined in [8], only 670 videos from this dataset are utilized for testing purposes.

**Evaluation Metrics.** For fair evaluation, this paper employs five widely used evaluation metrics in video captioning tasks to assess the quality of the generated captions. These metrics include BLEU@N[11], METEOR[12], ROUGE-L[13] and CIDEr[14]. Specifically, BLEU@n is commonly used to measure the accuracy of generated captions by comparing the n-gram overlap between reference and generated sentences; a higher overlap indicates more accurate generated descriptions. METEOR, an improvement upon BLEU, calculates recall and precision weighted by unigrams to assess matching relationships between sequences, synonyms, word roots, prefixes, and meanings between sentences. ROUGE-L is often used to evaluate the quality of text summarization by utilizing the longest common substring between sentences to compute precision and recall. CIDEr performs TF-IDF weighting on each n-gram between sentences and calculates the cosine similarity between their TF-IDF weighted vectors to measure the consistency between reference and generated descriptions, evaluating the consistency and richness of image descriptions.

**Implementation Details.** Our model adheres to an end-to-end training approach, where we solely input the raw features of both videos and sentences. Video features are extracted using Video Swin Transformer[15], employing the base-224 version, while textual features are extracted using Bert[16]. The parameters for Video Swin Transformer are based on the default settings, and for Bert, the maximum input token count is set to 24. The dimensions of video and text are both 768, and the dimensions of information content codebook is 64. During the training process, we conduct training on two V100 GPUs, with a batch size of 24 on each card. The initial learning rate is set to 1e-5. The warm-up iterations for training information content bias extractor is set to 6000 for MSVD, and 14000 for MSRVTT. $\tau_1$, $\tau_2$, $\tau_3$ are set to 0.2, 1, and 0.07, respectively.

Table 1 Performance comparison with SOTA methods on MSR-VTT and MSVD. All metrics indicate better performance when they having a higher numerical value.

| Methods | MSRVTT | | | | | MSVD | | | | |
|---|---|---|---|---|---|---|---|---|---|---|
| | Bleu@1 | Bleu@4 | Meteor | Rouge-L | CIDEr | Bleu@1 | Bleu@4 | Meteor | Rouge-L | CIDEr |
| UniVL | 80.51 | 41.79 | 28.94 | 60.78 | 50.04 | - | - | - | - | - |
| HMN | - | 43.5 | 29.0 | 62.7 | 51.5 | - | 59.2 | 37.7 | 75.1 | 104.0 |
| SwinBert | - | - | - | - | 53.8 | - | - | - | - | 120.6 |
| CoCap | - | 43.1 | 29.8 | 62.7 | 56.2 | - | 55.9 | 39.9 | 76.8 | 113.0 |
| Lavender | 81.84 | 43.12 | 29.63 | 62.51 | 55.17 | 88.65 | 63.96 | **43.80** | 80.01 | 135.10 |
| Ours | **82.44** | **44.34** | **29.88** | **62.94** | **57.17** | **88.96** | **64.43** | 43.37 | **80.27** | **138.68** |

### 3.2. Comparison with State-of-the-Art Methods

**Compared Methods.** In this section, we compare our proposed method with the state-of-the-art video captioning methods to demonstrate the effectiveness of our model. UniVL[17] pre-trained its model on large-scale dataset, and propose some innovative object functions especially for generation task. HMN[18] using object features, 2D and 3D

CNN features with multiple alignment loss with entity, predicate and sentence embedding to align multimodal embedding in a hierarchical way. SwinBert[19] use the pre-trained video extractor parameter, and propose a dynamic attention mask to reduce computational cost in multimodal fusion phase. CoCap[20]

**Results Comparison.** Table1 compares our method with SOTA methods on the MSR-VTT test split and MSVD test split. From this table, the following results can be derived: our method exhibits significant advantages compared to traditional non-end-to-end training approaches. Among these metrics, CIDEr stands out as relatively crucial and comprehensive. In comparison to HMN, our method achieves a notable improvement of 5.67% on the MSRVTT dataset and a 34.33% improvement on the MSVD dataset. Additionally, when compared with the end-to-end pre-train and finetune method, our approach demonstrates a 3.6% increase in CIDEr value on the MSVD dataset and a 2.0% increase on the MSRVTT dataset, outperforming the baseline.

### 3.3. Ablation Study

Table 2 Ablation studies on investigating contribution of our proposed bidirectional and margin contrastive losses on MSVD test split

| Methods | B@1 | B@4 | M | R | C |
|---|---|---|---|---|---|
| Baseline | 88.65 | 63.96 | **43.80** | 80.01 | 135.10 |
| $+L_{mcl}, L_b$ | 82.50 | 50.98 | 36.99 | 72.92 | 97.56 |
| $+L_{bfcl}$ | 88.64 | 63.60 | 43.34 | 79.59 | 137.10 |
| $+L_{gmc}$ | **88.96** | **64.43** | 43.37 | **80.27** | **138.68** |

In this section, we conducted ablation experiments on the effectiveness of the proposed loss on the MSVD dataset. Specifically, we tested the effectiveness of each loss individually by only adding one loss at a time. The experimental results in Table 2 show that when using only a single loss, $L_{bfcl}$ achieves an improvement of 2.0 on CIDEr, and $L_{mcl}$ achieves an improvement of 1.X points on CIDEr. However, when using only a single loss, other metrics show either marginal growth or values lower than the baseline. Our analysis suggests that this could be attributed to our excessive emphasis on information content size during the calculation of bias scores, neglecting considerations for other attributes of sentences, such as fluency and the length of quoted strings. When using only $L_{bfcl}$, the model rapidly achieves performance convergence in the initial phase but fails to capture complete information for tail sentences. Using only $L_{mcl}$ results in the model primarily fitting the bias extractor during the initial phase. However, in subsequent stages, due to the lack of a well-established feature initialization method, the effectiveness diminishes.

Combining $L_{bfcl}$ and $L_{mcl}$ addresses these issues and leads to a more comprehensive solution.

### 4. Conclusions

In this paper, we introduced a novel video captioning method. Firstly, we integrated contrastive learning into the training process of video captioning. Secondly, we addressed the connection between the imbalance in occurrence frequency and the imbalance in information content among training samples in the dataset. We quantified and trained this connection through a statistical approach. Finally, we incorporated the quantified scores as margins into the contrastive learning process to optimize the model's training effectiveness on tail samples, leading to an overall improvement. The results on two public datasets demonstrate the effectiveness of our method. In future work, we plan to further explore the impact of imbalanced distributions in datasets on model performance.

### Acknowledgements

This paper is supported by Center for Future Media, University of Electronic Science and Technology of China.

### References

[1] Zhou, L., Zhou, Y., Corso, J. J., Socher, R., and Xiong, C., "End-to-end Dense Video Captioning with Masked Transformer", Proceedings of CVPR 2018 Conference, Utah, pp. 8739-8748, June, 2018.

[2] Gao, L., Guo, Z., Zhang, H., Xu, X., and Shen, H. T., "Video Captioning with Attention-Based LSTM and Semantic Consistency", IEEE Transactions on Multimedia, vol. 19, no. 9, pp. 2045-2055, 2017.

[3] Yang, A., Nagrani, A., Seo, P. H., Miech, A., Pont-Tuset, J., Laptev, I., and Schmid, C., "Vid2seq: Large-scale pretraining of a visual language model for dense video captioning", Proceedings of CVPR2023 Conference, Oxford, pp. 10714-10726, September, 2023.

[4] Chen, S., He, X., Guo, L., Zhu, X., Wang, W., Tang, J., and Liu, J., "Valor: Vision-audio-language omni-perception pretraining model and dataset", arXiv preprint arXiv:2304.08345, 2023.

[5] He, X., Chen, S., Ma, F., Huang, Z., Jin, X., Liu, Z., and Feng, J., "VLAB: Enhancing Video Language Pre-training by Feature Adapting and Blending", arXiv preprint arXiv:2305.13167, 2023.

[6] Wang, J., Yang, Z., Hu, X., Li, L., Lin, K., Gan, Z., and Wang, L., "Git: A generative image-to-text transformer

for vision and language", arXiv preprint arXiv:2205.14100, 2022.
[7] Miech, A., Zhukov, D., Alayrac, J. B., Tapaswi, M., Laptev, I., and Sivic, J., "Howto100m: Learning a text-video embedding by watching hundred million narrated video clips", Proceedings of ICCV2019 Conference, Seoul, pp. 2630-2640, October, 2019.
[8] Li L, Gan Z, Lin K, et al. "Lavender: Unifying video-language understanding as masked language modeling", Proceedings of CVPR2023 Conference, Oxford, pp. 23119-23129, September, 2023.
[9] Xu, J., Mei, T., Yao, T., and Rui, Y., "Msr-vtt: A large video description dataset for bridging video and language", Proceedings of CVPR2016 Conference, Las Vegas, pp. 5288-5296, June, 2016.
[10] Chen, D., and Dolan, W. B., "Collecting highly parallel data for paraphrase evaluation", Proceedings of ACL2011 Conference, pp. 190-200, 2011.
[11] Papineni K., Roukos S., Ward T., and Zhu WJ., "Bleu: a method for automatic evaluation of machine translation", Proceedings of ACL2002 Conference, pp.311-318, 2002.
[12] Denkowski M, and Lavie A., "Meteor universal: Language specific translation evaluation for any target language", Proceedings of ACL workshop2014 Conference, pp.376-380, 2014.
[13] Lin CY. Rouge, "A package for automatic evaluation of summaries", In: Text summarization branches out, pp. 74-81, 2004.
[14] Vedantam R, Lawrence Zitnick C, and Parikh D. Cider, "Consensus-based image description evaluation", Proceedings of CVPR 2015 Conference, Boston, pp.4566-4575, June, 2015.
[15] Liu, Z., Ning, J., Cao, Y., Wei, Y., Zhang, Z., Lin, S., and Hu, H., "Video swin transformer", Proceedings of CVPR2022 Conference, New Orleans, pp. 3202-3211, June 2022.
[16] Devlin, J., Chang, M. W., Lee, K., and Toutanova, K., "Bert: Pre-training of deep bidirectional transformers for language understanding", arXiv preprint arXiv:1810.04805, 2018.
[17] Luo, H., Ji, L., Shi, B., Huang, H., Duan, N., Li, T., and Zhou, M., "Univl: A unified video and language pre-training model for multimodal understanding and generation", arXiv preprint arXiv:2002.06353, 2020.
[18] Ye H, Li G, Qi Y, et al. "Hierarchical modular network for video captioning", Proceedings of the CVPR 2022, New Orleans, pp. 17939-17948, June 2022.
[19] Lin K, Li L, Lin C C, et al. "Swinbert: End-to-end transformers with sparse attention for video captioning", Proceedings of CVPR 2022 Conference, New Orleans, pp: 17949-17958, June 2022.
[20] Shen, Y., Gu, X., Xu, K., Fan, H., Wen, L., and Zhang, L., "Accurate and Fast Compressed Video Captioning", Proceedings of ICCV2023 Conference, Paris, pp. 15558-15567, October, 2023.